\documentclass[compsoc, a4paper, 10 pt, conference]{IEEEtran}

\usepackage{etoolbox}
\makeatletter
\patchcmd{\@makecaption}
  {\scshape}
  {}
  {}
  {}
\makeatother     

\usepackage[font=small,labelfont=bf]{caption}
\usepackage{tabularx}
\usepackage{graphicx}
\usepackage{subcaption}
\usepackage{url}
\usepackage{calc}
\usepackage{amssymb}
\usepackage{amstext}
\usepackage{amsmath}
\usepackage{bm}
\usepackage[square, numbers, round, sort]{natbib}

\usepackage{rotating}
\usepackage{makecell}
\usepackage{multirow}

\usepackage{amssymb}
\usepackage{pifont}
\newcommand{\cmark}{\ding{51}}%

\newcommand{\btb}{\begin{tabular}}
\newcommand{\etb}{\end{tabular}}

\newcommand*\rot{\rotatebox{90}}

\usepackage{xcolor}
\definecolor{dimgray}{rgb}{0.4 0.4 0.45}

\usepackage[]{hyperref}
\hypersetup{
    colorlinks=true,
    allcolors=dimgray,
    linkcolor=dimgray,
    filecolor=dimgray,      
    urlcolor=dimgray,
}

\usepackage[noabbrev, capitalise, nameinlink]{cleveref}

\usepackage{extdash}

\title{\LARGE \bf
Uninformed Students: Student--Teacher Anomaly Detection \\ with Discriminative Latent Embeddings 
}

\author{Paul Bergmann \hspace{1cm} Michael Fauser \hspace{1cm} David Sattlegger \hspace{1cm} Carsten Steger \vspace{0.5cm} \\
MVTec Software GmbH\\
{\tt\small www.mvtec.com}\\
{\tt\small \{paul.bergmann, fauser, sattlegger, steger\}@mvtec.com}
}

\begin{document}

\maketitle
\thispagestyle{plain}
\pagestyle{plain}

\begin{abstract}
We introduce a powerful student--teacher framework for the challenging problem of unsupervised anomaly detection and pixel-precise anomaly segmentation in high-resolution images. Student networks are trained to regress the output of a descriptive teacher network that was pretrained on a large dataset of patches from natural images. This circumvents the need for prior data annotation. Anomalies are detected when the outputs of the student networks differ from that of the teacher network. This happens when they fail to generalize outside the manifold of anomaly-free training data. The intrinsic uncertainty in the student networks is used as an additional scoring function that indicates anomalies. We compare our method to a large number of existing deep learning based methods for unsupervised anomaly detection. Our experiments demonstrate improvements over state-of-the-art methods on a number of real-world datasets, including the recently introduced MVTec Anomaly Detection dataset that was specifically designed to benchmark anomaly segmentation algorithms.
\end{abstract}

\section{Introduction}

Unsupervised pixel-precise segmentation of regions that appear anomalous or novel to a machine learning model is an important and challenging task in many domains of computer vision. In automated industrial inspection scenarios, it is often desirable to train models solely on a single class of anomaly-free images to segment defective regions during inference. In an active learning setting, regions that are detected as previously unknown by the current model can be included in the training set to improve the model's performance. 

Recently, efforts have been made to improve anomaly detection for one-class or multi-class classification \cite{vae_novelty_recon_probability, andrews2016transfer, Burlina_2019_CVPR, chalapathy2018anomaly, masana2018, roitberg_bmvc_informed_democracy, pmlr-v80-ruff18a}. However, these algorithms assume that anomalies manifest themselves in the form of images of an entirely different class and a simple binary image-level decision whether an image is anomalous or not must be made. Little work has been directed towards the development of methods that can segment anomalous regions that only differ in a very subtle way from the training data. Bergmann et al. \cite{bergmann2019mvtec} provide benchmarks for several state-of-the-art algorithms and identify a large room for improvement.

\begin{figure}[t]
    \centering
    \includegraphics[width=0.99\linewidth]{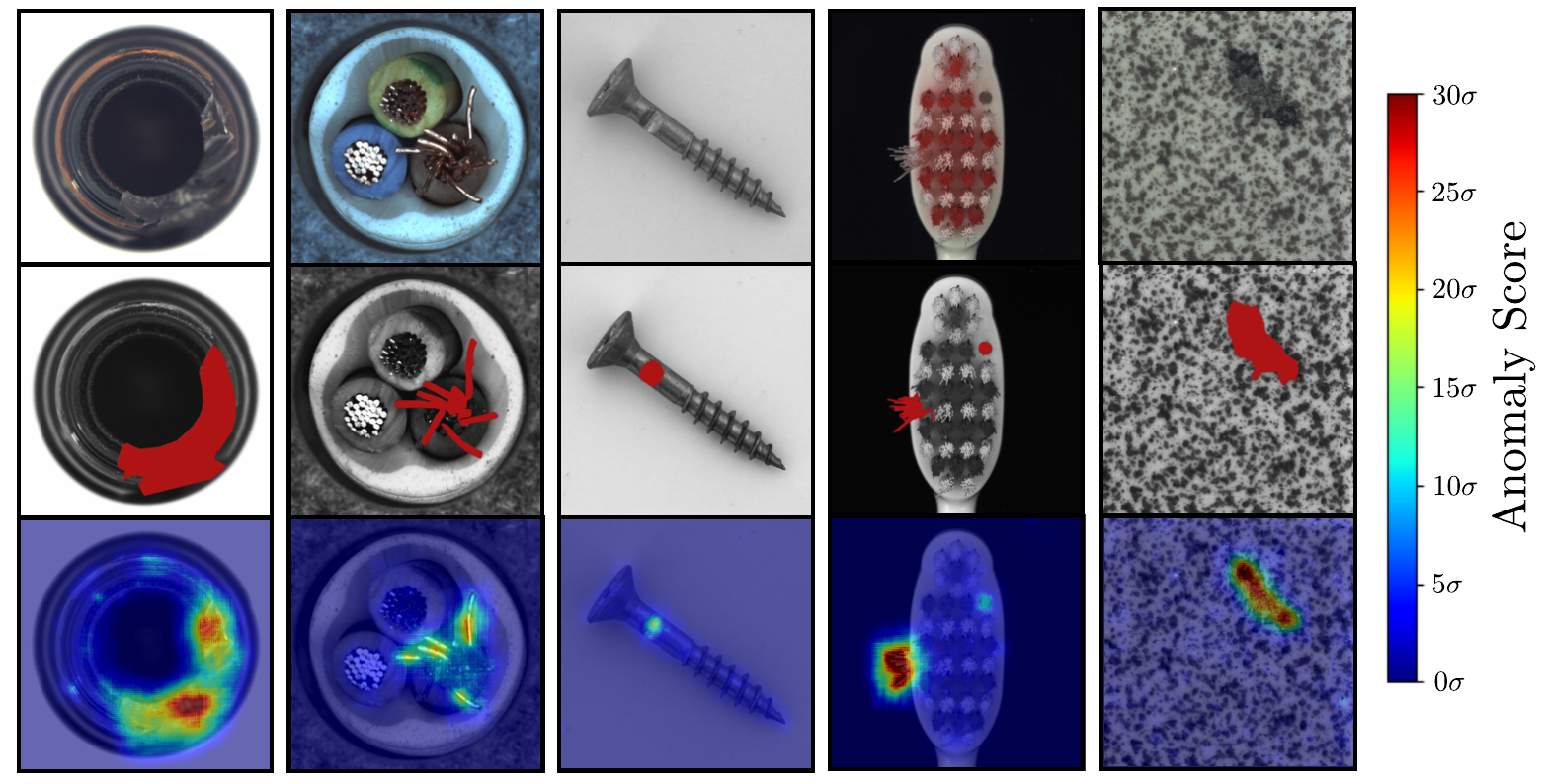}
    \caption{Qualitative results of our anomaly detection method on the MVTec Anomaly Detection dataset. \textbf{Top row:} Input images containing defects. \textbf{Center row:} Ground truth regions of defects in red. \textbf{Bottom row:} Anomaly scores for each image pixel predicted by our algorithm.}
    \label{fig:qual_results}
\end{figure}{}

\begin{figure*}[h!]
    \centering
    \includegraphics[width=0.95\linewidth]{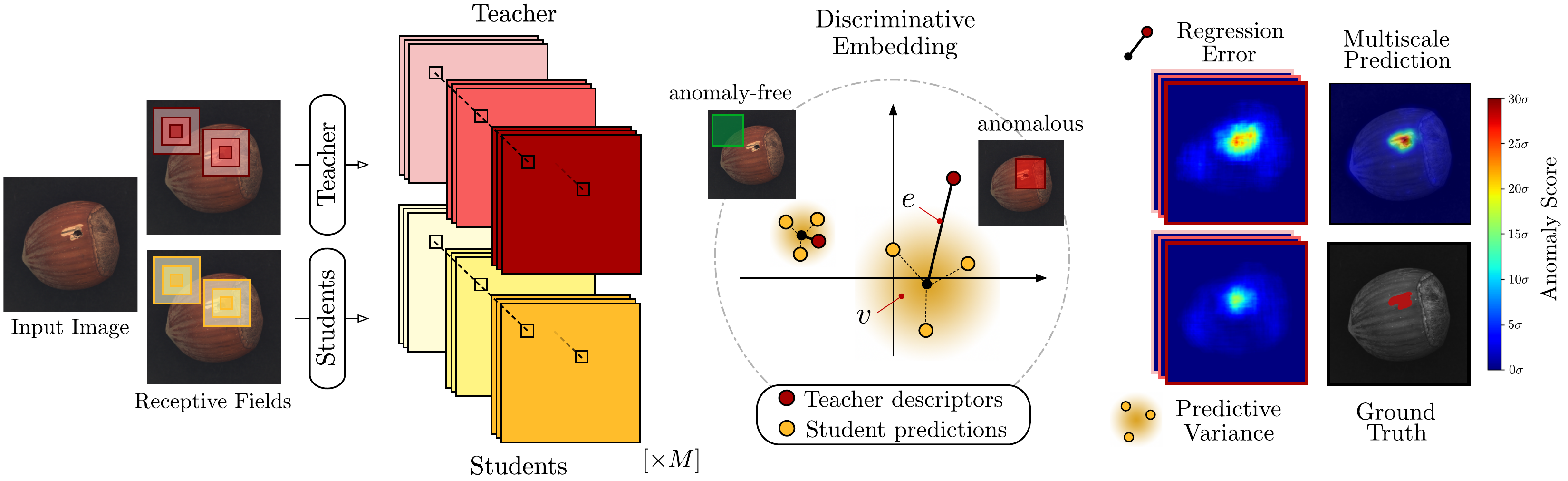}
    \caption{Schematic overview of our approach. Input images are fed through a teacher network that densely extracts features for local image regions. An ensemble of $M$ student networks is trained to regress the output of the teacher on anomaly-free data. During inference, the students will yield increased regression errors $e$ and predictive uncertainties $v$ in pixels for which the receptive field covers anomalous regions. Anomaly maps generated with different receptive fields can be combined for anomaly segmentation at multiple scales.}
    \label{fig:algorithm_overview}
\end{figure*}

Existing work predominantly focuses on generative algorithms such as Generative Adversarial Networks (GANs) \cite{schlegl2019_fast_anogan, schlegl_anogan} or Variational Autoencoders (VAEs) \cite{c_baur_vae_gan, q_space_golkov}. These detect anomalies using per-pixel reconstruction errors or by evaluating the density obtained from the model's probability distribution. This has been shown to be problematic due to inaccurate reconstructions or poorly calibrated likelihoods \cite{bergmann2018ssim, do_deep_gen_models_know_what_they_dont_know}.

The performance of many supervised computer vision algorithms \cite{Kornblith_2019_CVPR_transfer_learning_supervised_classification, Sun_2019_CVPR_transfer_learning_semseg} is improved by transfer learning, i.e.~by using discriminative embeddings from pretrained networks. For unsupervised anomaly detection, such approaches have not been thoroughly explored so far. Recent work suggests that these feature spaces generalize well for anomaly detection and even simple baselines outperform generative deep learning approaches \cite{Burlina_2019_CVPR, Perera_2019_CVPR_Transfer_Learning_Novelty_Detection}. However, the performance of existing methods on large high-resolution image datasets is hampered by the use of shallow machine learning pipelines that require a dimensionality reduction of the used feature space. Moreover, they rely on heavy training data subsampling since their capacity does not suffice to model highly complex data distributions with a large number of training samples. 

We propose to circumvent these limitations of shallow models by implicitly modeling the distribution of training features with a student--teacher approach. This leverages the high capacity of deep neural networks and frames anomaly detection as a feature regression problem. Given a descriptive feature extractor pretrained on a large dataset of patches from natural images (the teacher), we train an ensemble of student networks on anomaly-free training data to mimic the teacher's output. During inference, the students' predictive uncertainty together with their regression error with respect to the teacher are combined to yield dense anomaly scores for each input pixel. Our intuition is that students will generalize poorly outside the manifold of anomaly-free training data and start to make wrong predictions. Figure~\ref{fig:qual_results} shows qualitative results of our method when applied to images selected from the MVTec Anomaly Detection dataset \cite{bergmann2019mvtec}. A schematic overview of the entire anomaly detection process is given in Figure~\ref{fig:algorithm_overview}. 
Our main contributions are: 
\begin{itemize}
\setlength\itemsep{1em}

    \item We propose a novel framework for unsupervised anomaly detection based on student--teacher learning. Local descriptors from a pretrained teacher network serve as surrogate labels for an ensemble of students. Our models can be trained end-to-end on large unlabeled image datasets and make use of all available training data.

    \item We introduce scoring functions based on the students' predictive variance and regression error to obtain dense anomaly maps for the segmentation of anomalous regions in natural images. We describe how to extend our approach to segment anomalies at multiple scales by adapting the students' and teacher's receptive fields. 
    
    \item We demonstrate state-of-the-art performance on three real-world computer vision datasets. We compare our method to a number of shallow machine learning classifiers and deep generative models that are fitted directly to the teacher's feature distribution. We also compare it to recently introduced deep learning based methods for unsupervised anomaly segmentation.
    
\end{itemize}

\section{Related Work}

There exists an abundance of literature on anomaly detection \cite{pimentel2014review}. Deep learning based methods for the segmentation of anomalies strongly focus on generative models such as autoencoders \cite{autoencoder_autoregressive, bergmann2018ssim} or GANs \cite{schlegl_anogan}. These attempt to learn representations from scratch, leveraging no prior knowledge about the nature of natural images, and segment anomalies by comparing the input image to a reconstruction in pixel space. This can result in poor anomaly detection performance due to simple per-pixel comparisons or imperfect reconstructions \cite{bergmann2018ssim}.

\subsection{Anomaly Detection with Pretrained Networks}

Promising results have been achieved by transferring discriminative embedding vectors of pretrained networks to the task of anomaly detection by fitting shallow machine learning models to the features of anomaly-free training data. Andrews et al.\ \cite{andrews2016transfer} use activations from different layers of a pretrained VGG network and model the anomaly-free training distribution with a $\nu$-SVM. However, they only apply their method to image classification and do not consider the segmentation of anomalous regions. Similar experiments have been performed by Burlina et al.\ \cite{Burlina_2019_CVPR}. They report superior performance of discriminative embeddings compared to feature spaces obtained from generative models.

Nazare et al.\ \cite{nazare2018pre} investigate the performance of different off-the-shelf feature extractors pretrained on an image classification task for the segmentation of anomalies in surveillance videos. Their approach trains a 1-Nearest-Neighbor (1-NN) classifier on embedding vectors extracted from a large number of anomaly-free training patches. Prior to the training of the shallow classifier, the dimensionality of the network's activations is reduced using Principal Component Analysis (PCA). To obtain a spatial anomaly map during inference, the classifier must be evaluated for a large number of overlapping patches, which quickly becomes a performance bottleneck and results in rather coarse anomaly maps. Similarly, Napoletano et al.\ \cite{cnn_feature_dictionary_nanofibres} extract activations from a pretrained ResNet-18 for a large number of cropped training patches and model their distribution using K-Means clustering after prior dimensionality reduction with PCA\@. They also perform strided evaluation of test images during inference. Both approaches sample training patches from the input images and therefore do not make use of all possible training features. This is necessary since, in their framework, feature extraction is computationally expensive due to the use of very deep networks that output only a single descriptor per patch. Furthermore, since shallow models are employed for learning the feature distribution of anomaly-free patches, the available training information must be strongly reduced.

 \begin{figure}[t]
    \centering
    \includegraphics[width=0.97\linewidth]{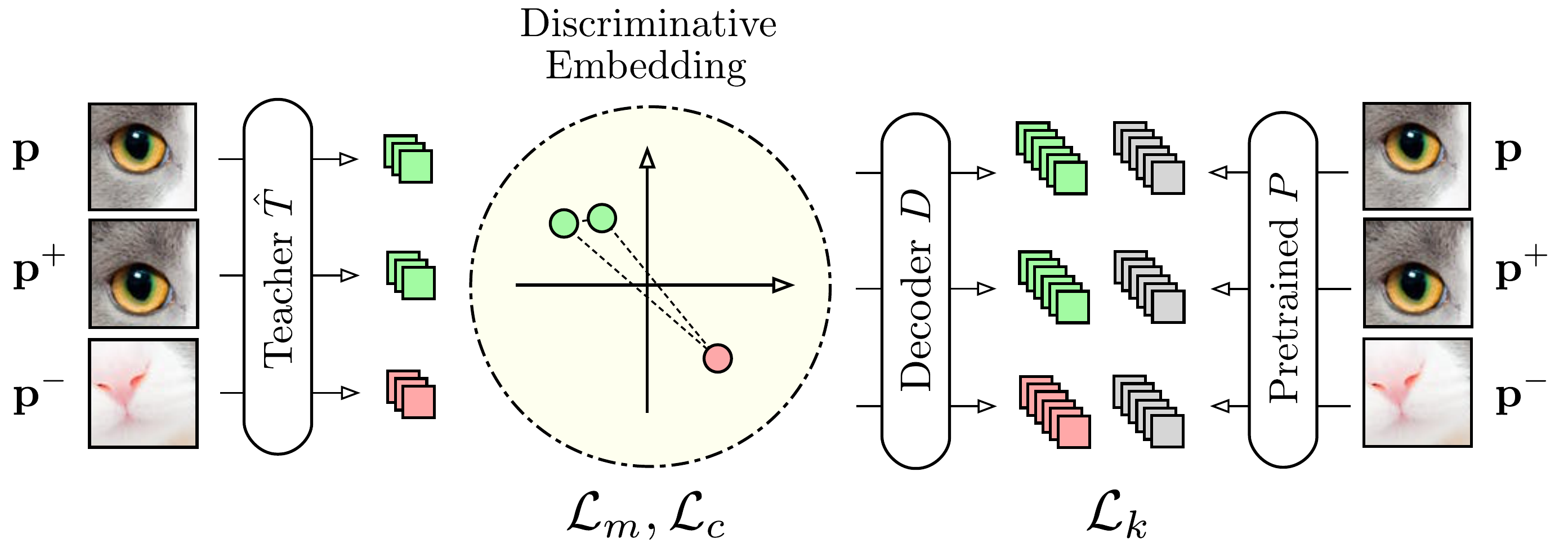}
    \caption{Pretraining of the teacher network $\hat{T}$ to output descriptive embedding vectors for patch-sized inputs. The knowledge of a powerful but computationally inefficient network $P$ is distilled into $\hat{T}$ by decoding the latent vectors to match the descriptors of $P$. We also experiment with embeddings obtained using self-supervised metric learning techniques based on triplet learning. Information within each feature dimension is maximized by decorrelating the feature dimensions within a minibatch.}
    \label{fig:pretraining}
\end{figure}{}

To circumvent the need for cropping patches and to speed up feature extraction, Sabokrou et al.\ \cite{sabokrou2018deep} extract descriptors from early feature maps of a pretrained AlexNet in a fully convolutional fashion and fit a unimodal Gaussian distribution to all available training vectors of anomaly-free images. Even though feature extraction is achieved more efficiently in their framework, pooling layers lead to a downsampling of the input image. This strongly decreases the resolution of the final anomaly map, especially when using descriptive features of deeper network layers with larger receptive fields. In addition, unimodal Gaussian distributions will fail to model the training feature distribution as soon as the problem complexity rises.

\subsection{Open-Set Recognition with Uncertainty Estimates}

Our work draws some inspiration from the recent success of open-set recognition in supervised settings such as image classification or semantic segmentation, where uncertainty estimates of deep neural networks have been exploited to detect out-of-distribution inputs using MC Dropout \cite{kendall2017uncertainties} or deep ensembles \cite{lakshminarayanan2017simple}. Seeboeck et al.\ \cite{seebock2019exploiting} demonstrate that uncertainties from segmentation networks trained with MC Dropout can be used to detect anomalies in retinal OCT images. Beluch et al.\ \cite{Beluch_2018_CVPR} show that the variance of network ensembles trained on an image classification task serves as an effective acquisition function for active learning. Inputs that appear anomalous to the current model are added to the training set to quickly enhance its performance. 

Such algorithms, however, demand prior labeling of images by domain experts for a supervised task, which is not always possible or desirable. In our work, we utilize feature vectors of pretrained networks as surrogate labels for the training of an ensemble of student networks. The predictive variance together with the regression error of the ensemble's output mixture distribution is then used as a scoring function to segment anomalous regions in test images.

\section{Student--Teacher Anomaly Detection}

This section describes the core principles of our proposed method. Given a training dataset $\mathcal{D} = \{\textbf{I}_1, \textbf{I}_2, \dots, \textbf{I}_N\}$ of anomaly-free images, our goal is to create an ensemble of \textit{student networks} $S_i$ that can later detect anomalies in test images $\textbf{J}$. This means that they can assign a score to each pixel indicating how much it deviates from the training data manifold. For this, the student models are trained against regression targets obtained from a descriptive \textit{teacher network} $T$ pretrained on a large dataset of natural images. After the training, anomaly scores can be derived for each image pixel from the students' regression error and predictive variance. Given an input image $\textbf{I} \in \mathbb{R}^{w \times h \times C}$ of width $w$, height $h$, and number of channels $C$, each student $S_i$ in the ensemble outputs a feature map $S_i(\textbf{I}) \in \mathbb{R}^{w \times h \times d}$. It contains descriptors $\textbf{y}_{(r,c)} \in \mathbb{R}^{d}$ of dimension $d$ for each input image pixel at row $r$ and column $c$. By design, we limit the students' receptive field, such that $\textbf{y}_{(r,c)}$ describes a square local image region $\textbf{p}_{(r,c)}$ of $\textbf{I}$ centered at $(r,c)$ of side length $p$. The teacher $T$ has the same network architecture as the student networks. However, it remains constant and extracts descriptive embedding vectors for each pixel of the input image $\textbf{I}$ that serve as deterministic regression targets during student training.

\begin{figure}[t]
    \centering
    \includegraphics[width=0.98\linewidth]{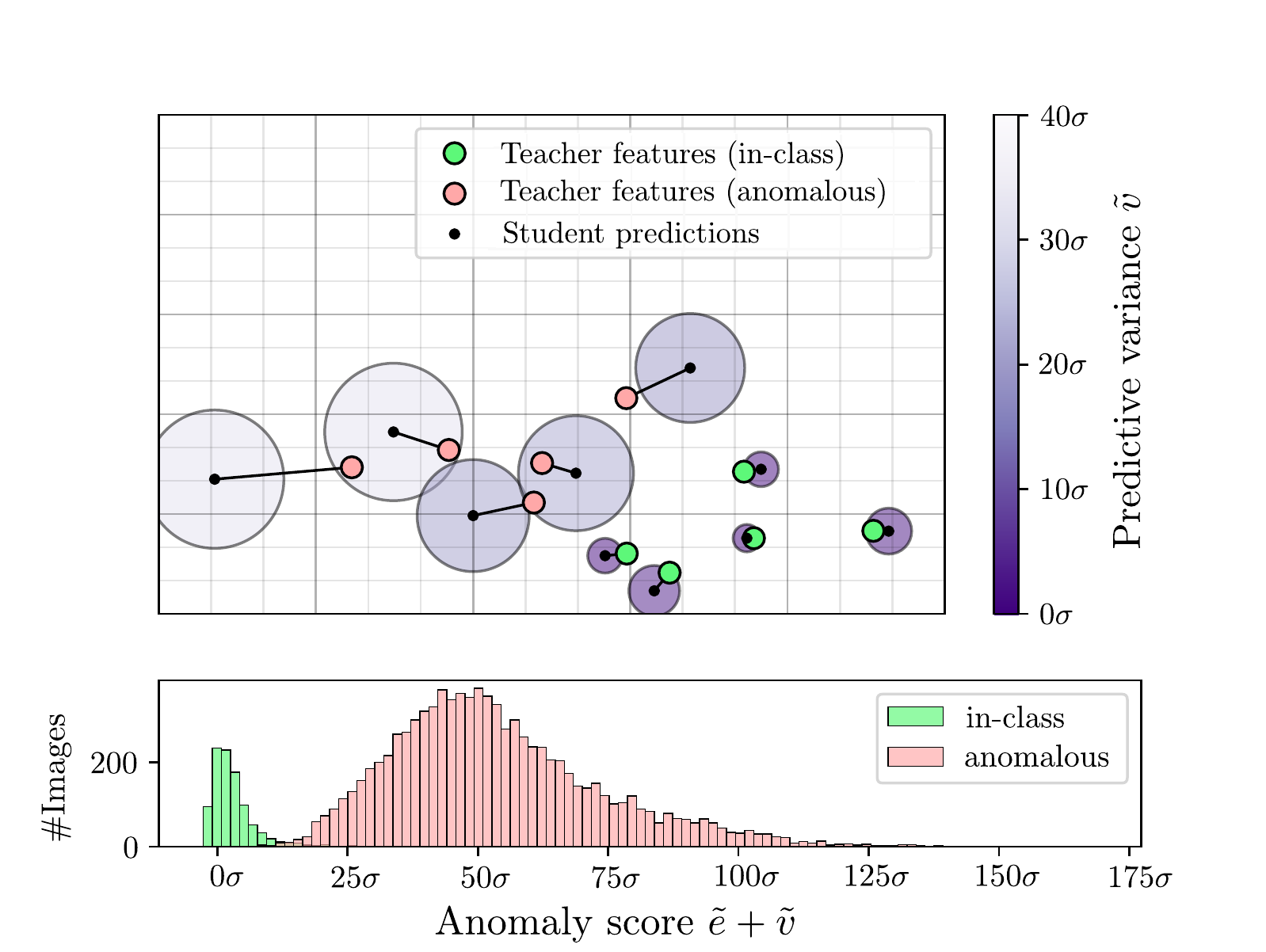}
    \caption{Embedding vectors visualized for ten samples of the MNIST dataset. Larger circles around the students' mean predictions indicate increased predictive variance. Being only trained on a single class of training images, the students manage to accurately regress the features solely for this class (green). They yield large regression errors and predictive uncertainties for images of other classes (red). Anomaly scores for the entire dataset are displayed in the bottom histogram.}
    \label{fig:example_result_mnist}
\end{figure}{}

\subsection{Learning Local Patch Descriptors}

We begin by describing how to efficiently construct a descriptive teacher network $T$ using metric learning and knowledge distillation techniques. In existing work for anomaly detection with pretrained networks, feature extractors only output single feature vectors for patch-sized inputs or spatially heavily downsampled feature maps \cite{cnn_feature_dictionary_nanofibres, sabokrou2018deep}. In contrast, our teacher network $T$ efficiently outputs descriptors for every possible square of side length $p$ within the input image. $T$ is obtained by first training a network $\hat{T}$ to embed patch-sized images $\textbf{p} \in \mathbb{R}^{p \times p \times C}$ into a metric space of dimension $d$ using only convolution and max-pooling layers. Fast dense local feature extraction for an entire input image can then be achieved by a deterministic network transformation of $\hat{T}$ to $T$ as described in \cite{stricker2017}. This yields significant speedups compared to previously introduced methods that perform patch-based strided evaluations. To let $\hat{T}$ output semantically strong descriptors, we investigate both self-supervised metric learning techniques as well as distilling knowledge from a descriptive but computationally inefficient pretrained network. A large number of training patches $\textbf{p}$ can be obtained by random crops from any image database. Here, we use ImageNet \cite{krizhevsky2012imagenet}.

\subsubsection*{Knowledge Distillation}

Patch descriptors obtained from deep layers of CNNs trained on image classification tasks perform well for anomaly detection when modeling their distribution with shallow machine learning models \cite{cnn_feature_dictionary_nanofibres, nazare2018pre}. However, the architectures of such CNNs are usually highly complex and computationally inefficient for the extraction of local patch descriptors. Therefore, we distill the knowledge of a powerful pretrained network $P$ into $\hat{T}$ by matching the output of $P$ with a decoded version of the descriptor obtained from $\hat{T}$:
\begin{equation}
  \mathcal{L}_{k}(\hat{T}) = ||D(\hat{T}(\textbf{p})) - P(\textbf{p})||^2.
\end{equation}
$D$ denotes a fully connected network that decodes the $d$-dimensional output of $\hat{T}$ to the output dimension of the pretrained network's descriptor.

\subsubsection*{Metric Learning}

If for some reason pretrained networks are unavailable, one can also learn local image descriptors in a fully self-supervised way \cite{unsupervised_patch_learning}. Here, we investigate the performance of discriminative embeddings obtained using triplet learning. For every randomly cropped patch $\textbf{p}$, a triplet of patches $(\textbf{p}, \textbf{p}^{+}, \textbf{p}^{-})$ is augmented. Positive patches $\textbf{p}^{+}$ are obtained by small random translations around $\textbf{p}$, changes in image luminance, and the addition of Gaussian noise. The negative patch $\textbf{p}^{-}$ is created by a random crop from a randomly chosen different image. In-triplet hard negative mining with anchor swap \cite{learning_local_image_descriptors_bmvc} is used as a loss function for learning an embedding sensitive to the $\ell_2$ metric
\begin{equation}
      \mathcal{L}_{m}(\hat{T}) = \max\{0, \delta + \delta^{+} - \delta^{-}\},
\end{equation}
where $\delta > 0$ denotes the margin parameter and in-triplet distances $\delta^+$ and $\delta^-$ are defined as:
\begin{align}
\delta^{+} &= ||\hat{T}(\textbf{p}) - \hat{T}(\textbf{p}^{+})||^2 \\
\delta^{-} &= \min\{||\hat{T}(\textbf{p}) - \hat{T}(\textbf{p}^{-})||^2, ||\hat{T}(\textbf{p}^{+}) - \hat{T}(\textbf{p}^{-})||^2\}.
\end{align}

\subsubsection*{Descriptor Compactness}

As proposed by Vassileios et al.\ \cite{l2_net}, we minimize the correlation between descriptors within one minibatch of inputs $\textbf{p}$ to increase the descriptors' compactness and remove unnecessary redundancy:
\begin{equation}
     \mathcal{L}_{c}(\hat{T}) = \sum_{i \neq j} c_{ij},
\end{equation}
where $c_{ij}$ denotes the entries of the correlation matrix computed over all descriptors $\hat{T}(\textbf{p})$ in the current minibatch. 

The final training loss for $\hat{T}$ is then given as
\begin{equation}
    \mathcal{L}(\hat{T}) = \lambda_{k}\mathcal{L}_{k}(\hat{T}) + 
  \lambda_{m}\mathcal{L}_{m}(\hat{T}) +
  \lambda_{c}\mathcal{L}_{c}(\hat{T}),
\end{equation}
where $\lambda_{k}$, $\lambda_{m}$, $\lambda_{c} \geq 0$ are weighting factors for the individual loss terms. Figure~\ref{fig:pretraining} summarizes the entire learning process for the teacher's discriminative embedding.

\subsection{Ensemble of Student Networks for Deep Anomaly Detection}

Next, we describe how to train student networks $S_i$ to predict the teacher's output on anomaly-free training data. We then derive anomaly scores from the students' predictive uncertainty and regression error during inference. First, the vector of component-wise means $\bm{\mu} \in \mathbb{R}^d$ and standard deviations $\bm{\sigma} \in \mathbb{R}^d$ over all training descriptors is computed for data normalization. Descriptors are extracted by applying $T$ to each image in the dataset $\mathcal{D}$. We then train an ensemble of $M \geq 1$ randomly initialized student networks $S_i$, $i \in \{1, \dots, M\}$ that possess the identical network architecture as the teacher $T$. For an input image $\textbf{I}$, each student outputs its predictive distribution over the space of possible regression targets for each local image region $\textbf{p}_{(r,c)}$ centered at row $r$ and column $c$. Note that the students' architecture with limited receptive field of size $p$ allows us to obtain dense predictions for each image pixel with only a single forward pass, without having to actually crop the patches $\textbf{p}_{(r,c)}$. The students' output vectors are modeled as a Gaussian distribution $Pr(\textbf{y}|\textbf{p}_{(r,c)}) = \mathcal{N}(\textbf{y} | \bm{\mu}^{S_i}_{(r,c)}, s)$ with constant covariance $s \in \mathbb{R}$, where $\bm{\mu}^{S_i}_{(r,c)}$ denotes the prediction made by $S_i$ for the pixel at $(r,c)$. Let $\textbf{y}_{(r,c)}^{T}$ denote the teacher's respective descriptor that is to be predicted by the students. The log-likelihood training criterion $\mathcal{L}(S_i)$ for each student network then simplifies to the squared $\ell_2$-distance in feature space:

\begin{table*}[t]
\centering
\resizebox{0.97\linewidth}{!}{%
\btb{ccccccccccc} 
\hline
 & Category & \btb[c]{@{}c@{}} Ours \\ $p=65$\etb & \btb[c]{@{}c@{}} 1-NN\etb & \btb[c]{@{}c@{}} OC-SVM\etb & \btb[c]{@{}c@{}} K-Means\etb & \btb[c]{@{}c@{}} $\ell_2$-AE\etb & \btb[c]{@{}c@{}} VAE\etb & \btb[c]{@{}c@{}} SSIM-AE\etb & \btb[c]{@{}c@{}} AnoGAN\etb & \btb[c]{@{}c@{}} CNN-Feature \\ Dictionary\etb \\
\hline
\multirow{5}{*}{\rot{Textures}} %
            & Carpet 		& \btb[c]{@{}c@{}} \textbf{0.695} \etb & \btb[c]{@{}c@{}} 0.512 \etb & \btb[c]{@{}c@{}} 0.355 \etb & \btb[c]{@{}c@{}} 0.253 \etb & \btb[c]{@{}c@{}} 0.456 \etb & \btb[c]{@{}c@{}} 0.501 \etb & \btb[c]{@{}c@{}} 0.647 \etb & \btb[c]{@{}c@{}} 0.204 \etb & \btb[c]{@{}c@{}} 0.469 \etb\\
\cline{2-11} & Grid 			& \btb[c]{@{}c@{}} 0.819 \etb & \btb[c]{@{}c@{}} 0.228 \etb & \btb[c]{@{}c@{}} 0.125 \etb & \btb[c]{@{}c@{}} 0.107 \etb & \btb[c]{@{}c@{}} 0.582 \etb & \btb[c]{@{}c@{}} 0.224 \etb & \btb[c]{@{}c@{}} \textbf{0.849} \etb & \btb[c]{@{}c@{}} 0.226 \etb & \btb[c]{@{}c@{}} 0.183 \etb\\
\cline{2-11} & Leather 		& \btb[c]{@{}c@{}} \textbf{0.819} \etb & \btb[c]{@{}c@{}} 0.446 \etb & \btb[c]{@{}c@{}} 0.306 \etb & \btb[c]{@{}c@{}} 0.308 \etb & \btb[c]{@{}c@{}} \textbf{0.819} \etb & \btb[c]{@{}c@{}} 0.635 \etb & \btb[c]{@{}c@{}} 0.561 \etb & \btb[c]{@{}c@{}} 0.378 \etb & \btb[c]{@{}c@{}} 0.641 \etb\\
\cline{2-11} & Tile 			& \btb[c]{@{}c@{}} \textbf{0.912} \etb & \btb[c]{@{}c@{}} 0.822  \etb & \btb[c]{@{}c@{}} 0.722\etb & \btb[c]{@{}c@{}} 0.779 \etb & \btb[c]{@{}c@{}} 0.897 \etb & \btb[c]{@{}c@{}} 0.870 \etb & \btb[c]{@{}c@{}} 0.175 \etb & \btb[c]{@{}c@{}} 0.177 \etb & \btb[c]{@{}c@{}} 0.797 \etb\\
\cline{2-11} & Wood 			& \btb[c]{@{}c@{}} 0.725 \etb & \btb[c]{@{}c@{}} 0.502 \etb & \btb[c]{@{}c@{}} 0.336 \etb & \btb[c]{@{}c@{}} 0.411 \etb & \btb[c]{@{}c@{}} \textbf{0.727} \etb & \btb[c]{@{}c@{}} 0.628 \etb & \btb[c]{@{}c@{}} 0.605 \etb & \btb[c]{@{}c@{}} 0.386 \etb & \btb[c]{@{}c@{}} 0.621 \etb\\
\hline
\multirow{10}{*}{\rot{Objects}} %
            & Bottle 		& \btb[c]{@{}c@{}} \textbf{0.918} \etb & \btb[c]{@{}c@{}} 0.898 \etb & \btb[c]{@{}c@{}} 0.850 \etb & \btb[c]{@{}c@{}} 0.495 \etb & 0.910   & \btb[c]{@{}c@{}} 0.897 \etb & \btb[c]{@{}c@{}} 0.834 \etb & \btb[c]{@{}c@{}} 0.620 \etb & \btb[c]{@{}c@{}} 0.742 \etb\\
\cline{2-11}	& Cable 		& \btb[c]{@{}c@{}} \textbf{0.865} \etb & \btb[c]{@{}c@{}} 0.806 \etb & \btb[c]{@{}c@{}} 0.431 \etb & \btb[c]{@{}c@{}} 0.513 \etb & 0.825   & 0.654                                 & \btb[c]{@{}c@{}} 0.478 \etb & \btb[c]{@{}c@{}} 0.383 \etb & \btb[c]{@{}c@{}} 0.558 \etb\\
\cline{2-11} & Capsule 		& \btb[c]{@{}c@{}} \textbf{0.916} \etb & \btb[c]{@{}c@{}} 0.631 \etb & \btb[c]{@{}c@{}} 0.554 \etb & \btb[c]{@{}c@{}} 0.387 \etb & 0.862   & \btb[c]{@{}c@{}} 0.526 \etb & \btb[c]{@{}c@{}} 0.860 \etb & \btb[c]{@{}c@{}} 0.306 \etb & \btb[c]{@{}c@{}} 0.306 \etb \\
\cline{2-11} & Hazelnut 		& \btb[c]{@{}c@{}} \textbf{0.937} \etb & \btb[c]{@{}c@{}} 0.861 \etb & \btb[c]{@{}c@{}} 0.616 \etb & \btb[c]{@{}c@{}} 0.698 \etb & 0.917   & 0.878  & \btb[c]{@{}c@{}} 0.916 \etb & \btb[c]{@{}c@{}} 0.698 \etb & \btb[c]{@{}c@{}} 0.844 \etb                               \\
\cline{2-11} & Metal nut 	& \btb[c]{@{}c@{}} \textbf{0.895} \etb & \btb[c]{@{}c@{}} 0.705 \etb & \btb[c]{@{}c@{}} 0.319 \etb & \btb[c]{@{}c@{}} 0.351 \etb & 0.830   & \btb[c]{@{}c@{}} 0.576 \etb & \btb[c]{@{}c@{}} 0.603 \etb & \btb[c]{@{}c@{}} 0.320 \etb & \btb[c]{@{}c@{}} 0.358 \etb\\
\cline{2-11} & Pill 			& \btb[c]{@{}c@{}} \textbf{0.935} \etb & \btb[c]{@{}c@{}} 0.725 \etb & \btb[c]{@{}c@{}} 0.544 \etb & \btb[c]{@{}c@{}} 0.514 \etb & 0.893   & \btb[c]{@{}c@{}} 0.769 \etb & \btb[c]{@{}c@{}} 0.830 \etb & \btb[c]{@{}c@{}} 0.776 \etb & \btb[c]{@{}c@{}} 0.460 \etb\\
\cline{2-11} & Screw 		& \btb[c]{@{}c@{}} \textbf{0.928} \etb & \btb[c]{@{}c@{}} 0.604 \etb & \btb[c]{@{}c@{}} 0.644 \etb & \btb[c]{@{}c@{}} 0.550 \etb & 0.754   & \btb[c]{@{}c@{}} 0.559 \etb & \btb[c]{@{}c@{}} 0.887 \etb & \btb[c]{@{}c@{}} 0.466 \etb & \btb[c]{@{}c@{}} 0.277 \etb\\
\cline{2-11} & Toothbrush 	& \btb[c]{@{}c@{}} \textbf{0.863} \etb & \btb[c]{@{}c@{}} 0.675 \\ \etb & \btb[c]{@{}c@{}} 0.538 \etb & \btb[c]{@{}c@{}} 0.337 \etb & 0.822  & \btb[c]{@{}c@{}} 0.693 \etb & \btb[c]{@{}c@{}} 0.784 \etb & \btb[c]{@{}c@{}} 0.749 \etb & \btb[c]{@{}c@{}} 0.151 \etb\\
\cline{2-11} & Transistor 	& \btb[c]{@{}c@{}} 0.701 \etb & \btb[c]{@{}c@{}} 0.680 \etb & \btb[c]{@{}c@{}} 0.496 \etb & \btb[c]{@{}c@{}} 0.399 \etb & \textbf{0.728}   & 0.626                                  & \btb[c]{@{}c@{}} 0.725 \etb & \btb[c]{@{}c@{}} 0.549 \etb & \btb[c]{@{}c@{}} 0.628 \etb\\
\cline{2-11} & Zipper 		& \btb[c]{@{}c@{}} \textbf{0.933} \etb & \btb[c]{@{}c@{}} 0.512 \etb & \btb[c]{@{}c@{}} 0.355 \etb & \btb[c]{@{}c@{}} 0.253 \etb & 0.839  & 0.549  & \btb[c]{@{}c@{}} 0.665 \etb & \btb[c]{@{}c@{}} 0.467 \etb & \btb[c]{@{}c@{}} 0.703 \etb                              \\
\hline
\hline
 & Mean & \textbf{0.857} & 0.640 & 0.479 & 0.423 & 0.790 & 0.639 & 0.694 & 0.443 & 0.515 \\
\hline
\etb
}
\vspace{0.005cm}
\caption{Results on the MVTec Anomaly Detection dataset. For each dataset category, the normalized area under the PRO-curve up to an average false positive rate per-pixel of 30\% is given. It measures the average overlap of each ground-truth region with the predicted anomaly regions for multiple thresholds. The best-performing method for each dataset category is highlighted in boldface.}
\label{table:result_classification}
\end{table*}

\begin{equation}
    \mathcal{L}(S_i) = \frac{1}{wh} \sum_{(r,c)} ||\bm{\mu}_{(r,c)}^{S_i} - (\textbf{y}_{(r,c)}^{T} - \bm{\mu})\mathrm{diag}(\bm{\sigma})^{-1}||_2^2,
\end{equation}
where $\mathrm{diag}(\bm{\sigma})^{-1}$ denotes the inverse of the diagonal matrix filled with the values in $\bm{\sigma}$.

\subsubsection*{Scoring Functions for Anomaly Detection}

Having trained each student to convergence, a mixture of Gaussians can be obtained at each image pixel by equally weighting the ensemble's predictive distributions. From it, measures of anomaly can be obtained in two ways: First, we propose to compute the regression error of the mixture's mean $\bm{\mu}_{(r,c)}$ with respect to the teacher's surrogate label:
\begin{align}
  e_{(r,c)} &= ||\bm{\mu}_{(r,c)} - (\textbf{y}_{(r,c)}^{T} - \bm{\mu})\mathrm{diag}(\bm{\sigma})^{-1}||_2^2 \\
  &= \Bigl|\Bigl|\frac{1}{M} \sum_{i=1}^{M} \bm{\mu}^{S_i}_{(r,c)} - (\textbf{y}_{(r,c)}^{T} - \bm{\mu})\mathrm{diag}(\bm{\sigma})^{-1}\Bigr|\Bigr|_2^2
\end{align}
The intuition behind this score is that the student networks will fail to regress the teacher's output within anomalous regions during inference since the corresponding descriptors have not been observed during training. Note that $e_{(r,c)}$ is non-constant even for $M=1$, where only a single student is trained and anomaly scores can be efficiently obtained with only a single forward pass through the student and teacher network, respectively. 

As a second measure of anomaly, we compute for each pixel the predictive uncertainty of the Gaussian mixture as defined by Kendall et al.\ \cite{kendall2017uncertainties}, assuming that the student networks generalize similarly for anomaly-free regions and differently in regions that contain novel information unseen during training:
\begin{equation}
    v_{(r,c)} = \frac{1}{M}\sum_{i=1}^{M} ||\bm{\mu}_{(r,c)}^{S_i}||_2^2 - ||\bm{\mu}_{(r,c)}||_2^2.
\end{equation}

To combine the two scores, we compute the means $e_{\mu}, v_{\mu}$ and standard deviations $e_{\sigma}, v_{\sigma}$ of all $e_{(r,c)}$ and $v_{(r,c)}$, respectively, over a validation set of anomaly-free images. Summation of the normalized scores then yields the final anomaly score:
\begin{equation}
    \tilde{e}_{(r,c)} + \tilde{v}_{(r,c)} = \frac{e_{(r,c)} - e_{\mu}}{e_{\sigma}} + \frac{v_{(r,c)} - v_{\mu}}{v_{\sigma}}.
\end{equation}

Figure~\ref{fig:example_result_mnist} illustrates the basic principles of our anomaly detection method on the MNIST dataset, where images with label~0 were treated as the normal class and all other classes were treated as anomalous. Since the images of this dataset are very small, we extracted a single feature vector for each image using $\hat{T}$ and trained an ensemble of $M=5$ patch-sized students to regress the teacher's output. This results in a single anomaly score for each input image. Feature descriptors were embedded into 2D using multidimensional scaling \cite{multidimensional_scaling} to preserve their relative distances.

\subsection{Multi-Scale Anomaly Segmentation}

If an anomaly only covers a small part of the teacher's receptive field of size $p$, the extracted feature vector predominantly describes anomaly-free traits of the local image region. Consequently, the descriptor can be predicted well by the students and anomaly detection performance will decrease. One could tackle this problem by downsampling the input image. This would, however, lead to an undesirable loss in resolution of the output anomaly map. 

Our framework allows for explicit control over the size of the students' and teacher's receptive field $p$. Therefore, we can detect anomalies at various scales by training multiple student--teacher ensemble pairs with varying values of $p$. At each scale, an anomaly map with the same size as the input image is computed. Given $L$ student--teacher ensemble pairs with different receptive fields, the normalized anomaly scores $\tilde{e}^{(l)}_{(r,c)}$ and $\tilde{v}^{(l)}_{(r,c)}$ of each scale $l$ can be combined by simple averaging:
\begin{equation}
    \frac{1}{L} \sum_{l=1}^{L} \left(\tilde{e}^{(l)}_{(r,c)} + \tilde{v}^{(l)}_{(r,c)}\right).
\end{equation}

\begin{figure*}[t!]
    \centering
    \includegraphics[width=1.0\linewidth]{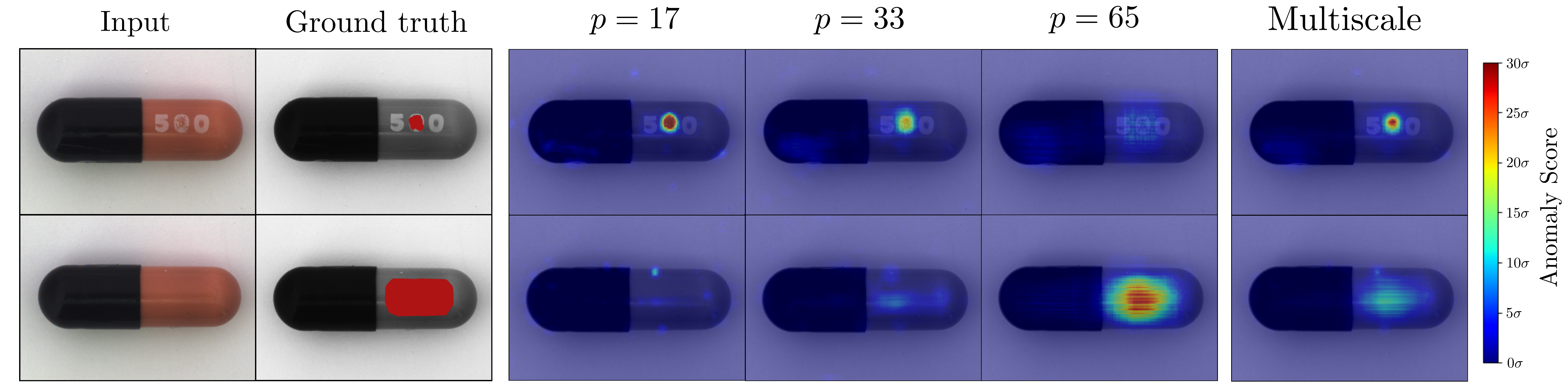}
    \caption{Anomaly detection at multiple scales: Architectures with receptive field of size $p=17$ manage to accurately segment the small scratch on the capsule (top row). However, defects at a larger scale such as the missing imprint (bottom row) become problematic. For increasingly larger receptive fields, the segmentation performance for the larger anomaly increases while it decreases for the smaller one. Our multiscale architecture mitigates this problem by combining multiple receptive fields.\vspace{0.1cm}}
    \label{fig:multiscale_benefit}
\end{figure*}{}

\section{Experiments}

To demonstrate the effectiveness of our approach, an extensive evaluation on a number of datasets is performed. We measure the performance of our student--teacher framework against existing pipelines that use shallow machine learning algorithms to model the feature distribution of pretrained networks. To do so, we compare to a K-Means classifier, a One-Class SVM (OC-SVM), and a 1-NN classifier. They are fitted to the distribution of the teacher's descriptors after prior dimensionality reduction using PCA\@. We also experiment with deterministic and variational autoencoders as deep distribution models over the teacher's discriminative embedding. The $\ell_2$-reconstruction error \cite{hadsell2006dimensionality} and reconstruction probability \cite{vae_novelty_recon_probability} are used as the anomaly score, respectively. We further compare our method to recently introduced generative and discriminative deep learning based anomaly detection models and report improved performance over the state of the art. We want to stress that the teacher has not observed images of the evaluated datasets during pretraining to avoid an unfair bias.

As a first experiment, we perform an ablation study to find suitable hyperparameters. Our algorithm is applied to a one-class classification setting on the MNIST \cite{mnist_dataset} and CIFAR-10 \cite{cifar_dataset} datasets. We then evaluate on the much more challenging MVTec Anomaly Detection (MVTec AD) dataset, which was specifically designed to benchmark algorithms for the segmentation of anomalous regions. It provides over 5000 high-resolution images divided into ten object and five texture categories. To highlight the benefit of our multi-scale approach, an additional ablation study is performed on MVTec AD, which investigates the impact of different receptive fields on the anomaly detection performance. 

For our experiments, we use identical network architectures for the student and teacher networks, with receptive field sizes $p \in \{17, 33, 65\}$. All architectures are simple CNNs with only convolutional and max-pooling layers, using leaky rectified linear units with slope $5 \times 10^{-3}$ as activation function. Table~\ref{table:ae_arch} shows the specific architecture used for $p=65$. For $p=17$ and $p=33$, similar architectures are given in in Appendix~\ref{AppendixA}. 

\begin{table}[b!]
\centering
\small
\def\arraystretch{1.1}
\begin{tabular}{clllcc}
\hline
Method               &    &         &       & MNIST & CIFAR-10                 \\ \hline
OCGAN \cite{Perera_2019_CVPR}               &    &         &      & 0.9750 & 0.6566               \\ \hline
1-NN                 &   &        & &    	0.9753    & 0.8189            \\ \hline
KMeans               &   &        & &    	0.9457  & 0.7592              \\ \hline
OC-SVM               &   &        & &    	0.9463   & 0.7388             \\ \hline
$\ell_2$-AE          &   &        &     & 	0.9832    & 0.7898            \\ \hline
VAE                  &   &        &     &	0.9535  & 0.7502              \\ \hline
\multicolumn{1}{l}{} & $L_{k}$ & $L_{m}$ & $L_{c}$ & \multicolumn{1}{l}{} & \multicolumn{1}{l}{} \\
Ours                 & \cmark  &        & \cmark     &	\textbf{0.9935}    & \textbf{0.8196}            \\ \hline
Ours                 & \cmark  & \cmark       & \cmark    	& 0.9926   & 0.8035             \\ \hline
Ours                 &   & \cmark       & \cmark    	& \textbf{0.9935}    & 0.7940            \\ \hline
Ours                 & \cmark  &        &     & 0.9917    & 0.8021            \\ \hline
\end{tabular}
\caption{Results on MNIST and CIFAR-10. For each method, the average area under the ROC curve is given, computed across each dataset category. For our algorithm, we evaluate teacher networks trained with different loss functions. \cmark\ corresponds to setting the respective loss weight to~$1$, otherwise it is set to~$0$.
}
\label{tab:results_mnist_cifar}
\end{table}

For the pretraining of the teacher networks $\hat{T}$, triplets augmented from the ImageNet dataset are used. Images are zoomed to equal width and height sampled from $\{4p, 4p+1, \dots, 16p\}$ and a patch of side length $p$ is cropped at a random location. A positive patch $\textbf{p}^+$ for each triplet is then constructed by randomly translating the crop location within the interval $\{-\frac{p-1}{4}, \dots, \frac{p-1}{4}\}$. Gaussian noise with standard deviation 0.1 is added to $\textbf{p}^+$. All images within a triplet are randomly converted to grayscale with a probability of 0.1. For knowledge distillation, we extract $512$-dimensional feature vectors from the fully connected layer of a ResNet-18 that was pretrained for classification on the ImageNet dataset. For network optimization, we use the Adam optimizer \cite{kingma2014adam} with an initial learning rate of $2 \times 10^{-4}$, a weight decay of $10^{-5}$, and a batch size of 64. Each teacher network outputs descriptors of dimension $d=128$ and is trained for $5 \times 10^{4}$ iterations.

\subsection{MNIST and CIFAR-10}

Before considering the problem of anomaly segmentation, we evaluate our method on the MNIST and CIFAR-10 datasets, adapted for one-class classification. Five students are trained on only a single class of the dataset, while during inference images of the other classes must be detected as anomalous. Each image is zoomed to the students' and teacher's input size $p$ and a single feature vector is extracted by passing it through the patch-sized networks $\hat{T}$ and $\hat{S}_i$. We examine different teacher networks by varying the weights $\lambda_{k}, \lambda_{m}, \lambda_{c}$ in the teacher's loss function $\mathcal{L}(\hat{T})$. The patch size for the experiments in this subsection is set to $p=33$. As a measure of anomaly detection performance, the area under the ROC curve is evaluated. Shallow and deep distributions models are trained on the teacher's descriptors of all available in-distribution samples. We additionally report numbers for OCGAN \cite{Perera_2019_CVPR}, a recently proposed generative model directly trained on the input images. Detailed information on training parameters for all methods on this dataset is found in Appendix~\ref{AppendixB}.

Table~\ref{tab:results_mnist_cifar} shows our results. Our approach outperforms the other methods for a variety of hyperparameter settings. Distilling the knowledge of the pretrained ResNet-18 into the teacher's descriptor yields slightly better performance than training the teacher in a fully self-supervised way using triplet learning. Reducing descriptor redundancy by minimizing the correlation matrix yields improved results. On average, shallow models and autoencoders fitted to our teacher's feature distribution outperform OCGAN but do not reach the performance of our approach. Since for 1-NN, every single training vector can be stored, it performs exceptionally well on these small datasets. On average, however, our method still outperforms all evaluated approaches.

\subsection{MVTec Anomaly Detection Dataset}

For all our experiments on MVTec AD, input images are zoomed to $w=h=256$ pixels. We train on anomaly-free images for 100 epochs with batch size 1. This is equivalent to training on a large number of patches per batch due to the limited size of the networks' receptive field.  We use Adam with initial learning rate $10^{-4}$ and weight decay $10^{-5}$. Teacher networks were trained with $\lambda_k = \lambda_c = 1$ and $\lambda_m = 0$, as this configuration performed best on MNIST and CIFAR-10. Ensembles contain $M=3$ students.

To train shallow classifiers on the teacher's output descriptors, a subset of vectors is randomly sampled from the teacher's feature maps. Their dimension is then reduced by PCA, retaining 95\% of the variance. The variational and deterministic autoencoders are implemented using a simple fully connected architecture and are trained on all available descriptors. In addition to fitting the models directly to the teacher's feature distribution, we benchmark our approach against the best performing deep learning based methods presented by Bergmann et al.\ \cite{bergmann2019mvtec} on this dataset. These methods include the CNN-Feature Dictionary \cite{cnn_feature_dictionary_nanofibres}, the SSIM-Autoencoder \cite{bergmann2018ssim}, and AnoGAN \cite{schlegl_anogan}. All hyperparameters are listed in detail in Appendix~\ref{AppendixC}.

\begin{table}[t!]
\centering
\small
\def\arraystretch{1.1}
\btb{cccccc} 
\hline
 & Category & \btb[c]{@{}c@{}} $p=17$\etb & \btb[c]{@{}c@{}} $p=33$\etb & \btb[c]{@{}c@{}} $p=65$\etb & \btb[c]{@{}c@{}} Multiscale\etb \\
\hline
\multirow{5}{*}{\rot{Textures}} %
            & Carpet 		& \btb[c]{@{}c@{}} 0.795 \etb & \btb[c]{@{}c@{}} \textbf{0.893} \etb & \btb[c]{@{}c@{}} 0.695 \etb &  0.879 \\
\cline{2-6} & Grid 			& \btb[c]{@{}c@{}} 0.920 \etb & \btb[c]{@{}c@{}} 0.949 \etb & \btb[c]{@{}c@{}} 0.819 \etb &  \textbf{0.952} \\
\cline{2-6} & Leather 		& \btb[c]{@{}c@{}} 0.935 \etb & \btb[c]{@{}c@{}} \textbf{0.956} \etb & \btb[c]{@{}c@{}} 0.819 \etb  & 0.945 \\
\cline{2-6} & Tile 			& \btb[c]{@{}c@{}} 0.936 \etb & \btb[c]{@{}c@{}} \textbf{0.950} \etb & \btb[c]{@{}c@{}} 0.912 \etb  & 0.946 \\
\cline{2-6} & Wood 			& \btb[c]{@{}c@{}} \textbf{0.943} \etb & \btb[c]{@{}c@{}} 0.929 \etb & \btb[c]{@{}c@{}} 0.725 \etb  & 0.911 \\
\hline
\multirow{10}{*}{\rot{Objects}} %
            & Bottle 		& \btb[c]{@{}c@{}} 0.814 \etb & \btb[c]{@{}c@{}} 0.890 \etb & \btb[c]{@{}c@{}} 0.918 \etb & \textbf{0.931} \\
\cline{2-6}	& Cable 		& \btb[c]{@{}c@{}} 0.671 \etb & \btb[c]{@{}c@{}} 0.764 \etb & \btb[c]{@{}c@{}} \textbf{0.865} \etb  & 0.818 \\
\cline{2-6} & Capsule 		& \btb[c]{@{}c@{}} 0.935 \etb & \btb[c]{@{}c@{}} 0.963 \etb & \btb[c]{@{}c@{}} 0.916 \etb & \textbf{0.968} \\
\cline{2-6} & Hazelnut 		& \btb[c]{@{}c@{}} 0.971 \etb & \btb[c]{@{}c@{}} \textbf{0.965} \etb & \btb[c]{@{}c@{}} 0.937 \etb & \textbf{0.965} \\
\cline{2-6} & Metal nut 	& \btb[c]{@{}c@{}} 0.891 \etb & \btb[c]{@{}c@{}} 0.928 \etb & \btb[c]{@{}c@{}} 0.895 \etb & \textbf{0.942} \\
\cline{2-6} & Pill 			& \btb[c]{@{}c@{}} 0.931 \etb & \btb[c]{@{}c@{}} 0.959 \etb & \btb[c]{@{}c@{}} 0.935 \etb & \textbf{0.961} \\
\cline{2-6} & Screw 		& \btb[c]{@{}c@{}} 0.915 \etb & \btb[c]{@{}c@{}} 0.937 \etb & \btb[c]{@{}c@{}} 0.928 \etb  & \textbf{0.942} \\
\cline{2-6} & Toothbrush 	& \btb[c]{@{}c@{}} \textbf{0.946} \etb & \btb[c]{@{}c@{}} 0.944 \etb & \btb[c]{@{}c@{}} 0.863 \etb  & 0.933 \\
\cline{2-6} & Transistor 	& \btb[c]{@{}c@{}} 0.540 \etb & \btb[c]{@{}c@{}} 0.611 \etb & \btb[c]{@{}c@{}} \textbf{0.701} \etb  & 0.666 \\
\cline{2-6} & Zipper 		& \btb[c]{@{}c@{}} 0.848 \etb & \btb[c]{@{}c@{}} 0.942 \etb & \btb[c]{@{}c@{}} 0.933 \etb & \textbf{0.951} \\
\hline
 & Mean & 0.866 & 0.900 & 0.857 & \textbf{0.914}  \\
\hline
\etb
\caption{Performance of our algorithm on the MVTec AD dataset for different receptive field sizes $p$. Combining anomaly scores across multiple receptive fields shows increased performance for many of the dataset's categories. We report the normalized area under the PRO curve up to an average false-positive rate of 30\%.\vspace{0.1cm}}
\label{table:result_hyperparameter}

\end{table}

We compute a threshold-independent evaluation metric based on the per-region-overlap (PRO), which weights ground-truth regions of different size equally. This is in contrast to simple per-pixel measures, such as ROC, for which a single large region that is segmented correctly can make up for many incorrectly segmented small ones. It was also used by Bergmann et al.\ in \cite{bergmann2019mvtec}. For computing the PRO metric, anomaly scores are first thresholded to make a binary decision for each pixel whether an anomaly is present or not. For each connected component within the ground truth, the relative overlap with the thresholded anomaly region is computed. We evaluate the PRO value for a large number of increasing thresholds until an average per-pixel false-positive rate of 30\% for the entire dataset is reached and use the area under the PRO curve as a measure of anomaly detection performance. Note that for high false-positive rates, large parts of the input images would be wrongly labeled as anomalous and even perfect PRO values would no longer be meaningful. We normalize the integrated area to a maximum achievable value of 1.

Table~\ref{table:result_classification} shows our results training each algorithm with a receptive field of $p=65$ for comparability. Our method consistently outperforms all other evaluated algorithms for almost every dataset category. The shallow machine learning algorithms fitted directly to the teacher's descriptors after applying PCA do not manage to perform satisfactorily for most of the dataset categories. This shows that their capacity does not suffice to accurately model the large number of available training samples. The same can be observed for the CNN-Feature Dictionary. As it was the case in our previous experiment on MNIST and CIFAR-10, 1-NN yields the best results amongst the shallow models.  Utilizing a large number of training features together with deterministic autoencoders increases the performance, but still does not match the performance of our approach. Current generative methods for anomaly segmentation such as Ano-GAN and the SSIM-autoencoder perform similar to the shallow methods fitted to the discriminative embedding of the teacher. This indicates that there is indeed a gap between methods that learn representations for anomaly detection from scratch and methods that leverage discriminative embeddings as prior knowledge.

\begin{table}[t!]
\centering
\small
\def\arraystretch{1.1}
\begin{tabular}{lcccc}
\hline
Layer & \multicolumn{1}{l}{Output Size} & \multicolumn{3}{c}{Parameters}                            \\
               & \multicolumn{1}{l}{}                     & Kernel      & Stride     \\ \hline
Input          & 65$\times$65$\times$3                                  & \multicolumn{1}{l}{} & \multicolumn{1}{l}{} & \multicolumn{1}{l}{} \\
Conv1          & 61$\times$61$\times$128                                & 5$\times$5                  & 1                    \\
MaxPool        & 30$\times$30$\times$128                                & 2$\times$2                  & 2                    \\
Conv2          & 26$\times$26$\times$128                                & 5$\times$5                  & 1                    \\
MaxPool        & 13$\times$13$\times$128                                & 2$\times$2                  & 2                    \\
Conv3          & 9$\times$9$\times$128                                  & 5$\times$5                  & 1                    \\
MaxPool        & 4$\times$4$\times$256                                  & 2$\times$2                  & 2                    \\
Conv4          & 1$\times$1$\times$256                                  & 4$\times$4                  & 1                    \\
Conv5          & 1$\times$1$\times$128                                  & 3$\times$3                  & 1                    \\
Decode         & 1$\times$1$\times$512                                  & 1$\times$1                  & 1                    \\ \hline
\end{tabular}
\caption{General outline of our network architecture for training teachers $\hat{T}$ with receptive field size $p=65$. Leaky rectified linear units with slope $5 \times 10^{-3}$ are applied as activation functions after each convolution layer. Architectures for $p=17$ and $p=33$ are given in Appendix~\ref{AppendixA}.\vspace{0.2cm}}
\label{table:ae_arch}
\end{table}

Table~\ref{table:result_hyperparameter} shows the performance of our algorithm for different receptive field sizes $p \in \{17, 33, 65\}$ and when combining multiple scales. For some objects, such as \textit{bottle} and \textit{cable}, larger receptive fields yield better results. For others, such as \textit{wood} and \textit{toothbrush}, the inverse behavior can be observed. Combining multiple scales enhances the performance for many of the dataset categories. A qualitative example highlighting the benefit of our multi-scale anomaly segmentation is visualized in Figure~\ref{fig:multiscale_benefit}.

\begin{table*}[t!]
\centering

\begin{subtable}{.5\textwidth}
\centering
\def\arraystretch{1.1}
\begin{tabular}{lcccc}
\hline
Layer & \multicolumn{1}{l}{Output Size} & \multicolumn{3}{c}{Parameters}                            \\
               & \multicolumn{1}{l}{}                     & Kernel      & Stride     \\ \hline
Input          & 33$\times$33$\times$3                                  & \multicolumn{1}{l}{} & \multicolumn{1}{l}{} & \multicolumn{1}{l}{} \\
Conv1          & 29$\times$29$\times$128                                & 3$\times$3                  & 1                    \\
MaxPool        & 14$\times$14$\times$128                                & 2$\times$2                  & 2                    \\
Conv2          & 10$\times$10$\times$256                                & 5$\times$5                  & 1                    \\
MaxPool        & 5$\times$5$\times$256                                & 2$\times$2                  & 2                    \\
Conv3          & 4$\times$4$\times$256                                  & 2$\times$2                  & 1                    \\
Conv4          & 1$\times$1$\times$128                                  & 4$\times$4                  & 1                    \\
Decode         & 1$\times$1$\times$512                                  & 1$\times$1                  & 1                    \\ \hline
\end{tabular}
\caption{Architecture for $p=33$.}
\label{table:arch_33}

\end{subtable}
\begin{subtable}{.5\textwidth}
\centering
\def\arraystretch{1.1}
\begin{tabular}{lcccc}
\hline
Layer & \multicolumn{1}{l}{Output Size} & \multicolumn{3}{c}{Parameters}                            \\
               & \multicolumn{1}{l}{}                     & Kernel      & Stride     \\ \hline
Input          & 17$\times$17$\times$3                                  & \multicolumn{1}{l}{} & \multicolumn{1}{l}{} & \multicolumn{1}{l}{} \\
Conv1          & 12$\times$12$\times$128                                & 5$\times$5                  & 1                    \\
Conv2        & 8$\times$8$\times$256                                & 5$\times$5                  & 1                    \\
Conv3          & 4$\times$4$\times$256                                & 5$\times$5                  & 1                    \\
Conv4        & 1$\times$1$\times$128                                & 4$\times$4                  & 1                    \\
Decode          & 1$\times$1$\times$512                                  & 1$\times$1                  & 1                    \\ \hline
\end{tabular}
\caption{Architecture for $p=17$.}
\label{table:arch_17}
\end{subtable}

\caption{Network architectures for teacher networks $\hat{T}$ with different receptive field sizes $p$.}

\end{table*}
\section{Conclusion}

We have proposed a novel framework for the challenging problem of unsupervised anomaly segmentation in natural images. Anomaly scores are derived from the predictive variance and regression error of an ensemble of student networks, trained against embedding vectors from a descriptive teacher network. Ensemble training can be performed end-to-end and purely on anomaly-free training data without requiring prior data annotation. Our approach can be easily extended to detect anomalies at multiple scales. We demonstrate improvements over current state-of-the-art methods on a number of real-world computer vision datasets for one-class classification and anomaly segmentation. 

\appendices

\section{Network Architectures}
\label{AppendixA}

A description of the network architecture for a patch-sized teacher network $\hat{T}$ with receptive field of size $p=65$ can be found in our main paper (Table 4). Architectures for teachers with receptive field sizes $p=33$ and $p=17$ are depicted in Tables~\ref{table:arch_33} and~\ref{table:arch_17}, respectively. Leaky rectified linear units with slope $5 \times 10^{-3}$ are used as activation function after each convolution layer.

\section{Experiments on MNIST and CIFAR-10}
\label{AppendixB}

Here, we give details about additional hyperparameters for our experiments on the MNIST and CIFAR-10 datasets. We additionally provide the per-class ROC-AUC values for the two datasets in Tables~\ref{table:mnist_results} and~\ref{table:cifar_results}, respectively.

\subsubsection*{Hyperparameter Settings}

 For the deterministic $\ell_2$-autoencoder ($\ell_2$-AE) and the variational autoencoder (VAE), we use a fully connected encoder architecture of shape 128--64--32--10 with leaky rectified linear units of slope $5 \times 10^{-3}$. The decoder is constructed in a manner symmetric to the encoder. Both autoencoders are trained for 100 epochs at an initial learning rate of $10^{-2}$ using the Adam optimizer and a batch size of 64. A weight decay rate of $10^{-5}$ is applied for regularization. To evaluate the reconstruction probability of the VAE, five independent forward passes are performed for each feature vector. For the One-Class SVM (OC-SVM), a radial basis function kernel is used. K-Means is trained with 10 cluster centers and the distance to the single closest cluster center is evaluated as the anomaly score for each input sample. For 1-NN, the feature vectors of all available training samples are stored and tested during inference. 

\section{Experiments on MVTec AD}
\label{AppendixC}

We give additional information on the hyperparameters used in our experiments on MVTec AD for both shallow machine learning models as well as deep learning methods.

\subsubsection*{Shallow Machine Learning Models}

For the 1-NN classifier, we construct a dictionary of 5000 feature vectors and take the distance to the closest training sample as anomaly score. For the other shallow classifiers, we fit their parameters on 50\,000 training samples, randomly chosen from the teacher's feature maps. The K-Means algorithm is run with 10 cluster centers and measures the distance to the nearest cluster center in the feature space during inference. The OC-SVM employs a radial basis function kernel.

\subsubsection*{Deep-Learning Based Models}

For evaluation on MVTec AD, the architecture of the $\ell_2$-AE and VAE are identical to the ones used on the MNIST and CIFAR-10 dataset. Each fully connected autoencoder is trained for 100 epochs. We use Adam with initial learning rate $10^{-4}$ and weight decay $10^{-5}$. Batches are constructed from 512 randomly sampled vectors of the teacher's feature maps. The reconstruction probability of the VAE is computed by five individual forward passes through the network. For the evaluation of AnoGAN, the SSIM-Autoencoder, and the CNN-Feature Dictionary, we use the same hyperparameters as Bergmann et al.\ in the MVTec AD dataset paper \cite{bergmann2019mvtec}. Only a slight adaption is applied to the CNN-Feature Dictionary by cropping patches of size $p=65$ and performing the evaluation by computing anomaly scores for overlapping patches with a stride of 4 pixels.

\subsubsection*{Qualitative Results}

\begin{figure*}

\centering
\includegraphics[width=0.9\textwidth]{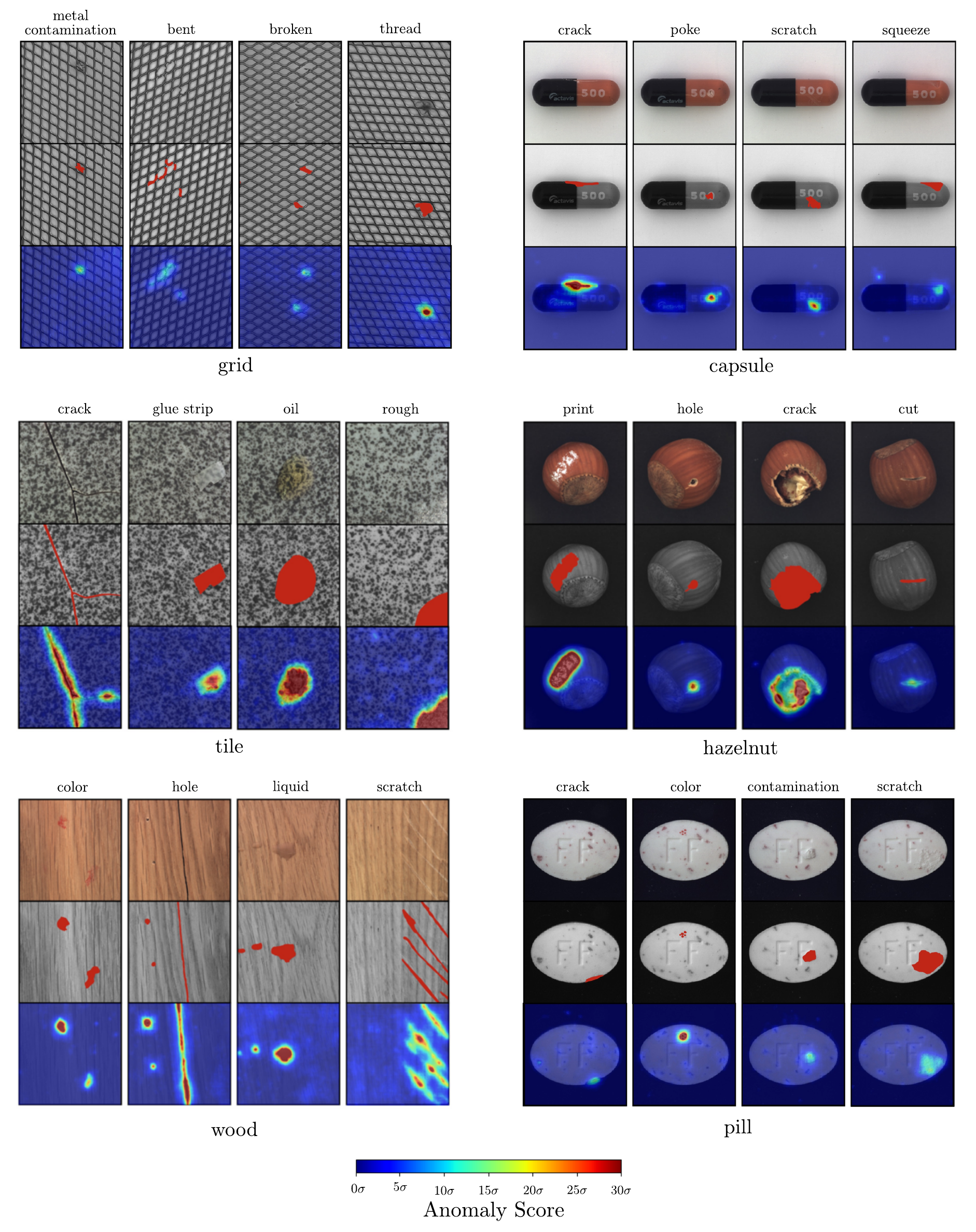}
\caption{Qualitative results of our method on selected textures (left) and objects (right) of the MVTec Anomaly Detection dataset. Our algorithm performs robustly across various defect categories, such as color defects, contaminations, and structural anomalies. \textbf{Top row:} Input images containing defects. \textbf{Center row:} Ground truth regions of defects in red. \textbf{Bottom row:} Anomaly scores for each image pixel predicted by our algorithm.}
\label{fig_qual}
\end{figure*}

We provide additional qualitative results of our method on MVTec AD for three objects and three textures in Figure \ref{fig_qual}. For each category, anomaly maps for multiple defect classes are provided. Our method performs well across different defect types and sizes. The results are shown for an ensemble of 3 students and a multi-scale architecture of receptive field sizes in $\{17, 33, 65\}$ pixels.

\begin{table*}[]
\centering
\resizebox{0.97\linewidth}{!}{%
\begin{tabular}{clllccccccccccc}
\hline
Method               &    &         &      & 0                    & 1                    & 2                    & 3                    & 4                    & 5                    & 6                    & 7                    & 8                    & 9                    & Mean                 \\ \hline
OCGAN               &    &         &      & 0.998	& \textbf{0.999}	& 0.942	& 0.963	& 0.975	& 0.980	& 0.991	& 0.981	& 0.939	& 0.981	& 0.9750                \\ \hline
1-NN                 &   &        &     & 0.989 &	0.998 &	0.962 &	0.970 &	0.980 &	0.955 &	0.979 &	0.981 &	0.968 &	0.971 &	0.9753                \\ \hline
KMeans                 &   &        &     & 0.973 &	0.995 &	0.898 &	0.948 &	0.960 &	0.920 &	0.948 &	0.948 &	0.940 &	0.927 &	0.9457                \\ \hline
OC-SVM                 &   &        &     & 0.980 &	0.998 &	0.887 &	0.944 &	0.964 &	0.909 &	0.949 &	0.957 &	0.935 &	0.940 &	0.9463                \\ \hline
$\ell_2$-AE                 &   &        &     & 0.992 &	\textbf{0.999} &	0.967 &	0.980 &	0.988 &	0.970 &	0.988 &	0.987 &	0.978 &	0.983 &	0.9832                \\ \hline
VAE                 &   &        &     & 0.983 &	0.998 &	0.915 &	0.941 &	0.969 &	0.925 &	0.964 &	0.940 &	0.955 &	0.945 &	0.9535                \\ \hline
\multicolumn{1}{l}{} & $L_{k}$ & $L_{m}$ & $L_{c}$ & \multicolumn{1}{l}{} & \multicolumn{1}{l}{} & \multicolumn{1}{l}{} & \multicolumn{1}{l}{} & \multicolumn{1}{l}{} & \multicolumn{1}{l}{} & \multicolumn{1}{l}{} & \multicolumn{1}{l}{} & \multicolumn{1}{l}{} & \multicolumn{1}{l}{} & \multicolumn{1}{l}{} \\
Ours                 & \cmark  &        & \cmark    & \textbf{0.999} &	\textbf{0.999} &	0.990 &	\textbf{0.993} & \textbf{0.992} &	\textbf{0.993} &	\textbf{0.997} &	\textbf{0.995} &	0.986 &	0.991 &	\textbf{0.9935}                \\ \hline
Ours                 & \cmark  & \cmark       & \cmark    & \textbf{0.999}	& \textbf{0.999}	& 0.988	& 0.992	& 0.988	& \textbf{0.993}	& \textbf{0.997}	& \textbf{0.995}	& 0.984	& 0.991	& 0.9926                \\ \hline
Ours                 &   & \cmark       & \cmark    & \textbf{0.999}	& \textbf{0.999}	& \textbf{0.992}	& 0.992	& 0.988	& \textbf{0.993}	& \textbf{0.997}	& \textbf{0.995}	& \textbf{0.988}	& \textbf{0.992}	& \textbf{0.9935}                \\ \hline
Ours                 & \cmark  &        &     & \textbf{0.999}	& \textbf{0.999}	& 0.989	& 0.990	& 0.990	& 0.990	& \textbf{0.997}	& 0.993	& 0.981	& 0.989	& 0.9917                \\ \hline
\end{tabular}
}
\caption{Results on the MNIST dataset. For each method and digit, the area under the ROC curve is given. For our algorithm, we evaluate teacher networks trained with different loss functions. \cmark\ corresponds to setting the respective loss weight to~1, otherwise it is set to~0.}
\label{table:mnist_results}
\end{table*}

\begin{table*}[]
\centering
\resizebox{0.97\linewidth}{!}{%
\begin{tabular}{clllccccccccccc}
\hline
Method               &    &         &      & airplane                    & automobile                    & bird                    & cat                    & deer                    & dog                    & frog                    & horse                    & ship                    & truck                    & Mean                 \\ \hline
OCGAN                &    &         &      & 0.757	& 0.531	& 0.640	& 0.620	& 0.723	& 0.620	& 0.723	& 0.575 & 0.820 &	0.554 &	0.6566                \\ \hline 
1-NN                 &   &        &     & 0.792 &	\textbf{0.860} &	\textbf{0.746} &	0.729 &	0.815 &	\textbf{0.797} &	0.876 &	\textbf{0.836} &	0.856 &	\textbf{0.882} &	0.8189                \\ \hline
KMeans                 &   &        &     & 0.673 &	0.822 &	0.665 &	0.676 &	0.742 &	0.746 &	0.828 &	0.780 &	0.817 &	0.843 &	0.7592                \\ \hline
OC-SVM                 &   &        &     & 0.651 &	0.785 &	0.618 &	0.679 &	0.733 &	0.730 &	0.797 &	0.760 &	0.799 &	0.836 &	0.7388                \\ \hline
$\ell_2$-AE                  &   &        &     & 0.747 &	0.862 &	0.690 &	0.698 &	0.788 &	0.759 &	0.849 &	0.824 &	0.812 &	0.869 &	0.7898                \\ \hline
VAE                 &   &        &     & 0.705 &	0.819 &	0.605 &	0.700 &	0.734 &	0.731 &	0.797 &	0.751 &	0.801 &	0.859 &	0.7502                \\ \hline
\multicolumn{1}{l}{} & $L_{k}$ & $L_{m}$ & $L_{c}$ & \multicolumn{1}{l}{} & \multicolumn{1}{l}{} & \multicolumn{1}{l}{} & \multicolumn{1}{l}{} & \multicolumn{1}{l}{} & \multicolumn{1}{l}{} & \multicolumn{1}{l}{} & \multicolumn{1}{l}{} & \multicolumn{1}{l}{} & \multicolumn{1}{l}{} & \multicolumn{1}{l}{} \\
Ours                 & \cmark  &        & \cmark    & 0.789 &	0.849 &	0.734 &	\textbf{0.748} &	\textbf{0.851} &	0.793 &	\textbf{0.892} &	0.830 &	\textbf{0.862} &	0.848 &	\textbf{0.8196}                \\ \hline
Ours                 & \cmark  & \cmark       & \cmark    & 0.784	& 0.836	& 0.706	& 0.742	& 0.826	& 0.768	& 0.870	& 0.815	& 0.857	& 0.831	& 0.8035                \\ \hline
Ours                 &   & \cmark       & \cmark    & \textbf{0.804}	& 0.855	& 0.706	& 0.709	& 0.798	& 0.738	& 0.860	& 0.797	& 0.849	& 0.824	& 0.7940                \\ \hline
Ours                 & \cmark  &        &     & 0.766	& 0.817	& 0.715	& 0.736	& 0.855	& 0.763	& 0.885	& 0.819	& 0.838	& 0.827	& 0.8021                \\ \hline
\end{tabular}
}
\caption{Results on the CIFAR-10 dataset. For each method and class, the area under the ROC curve is given. For our algorithm, we evaluate teacher networks trained with different loss functions. \cmark\ corresponds to setting the respective loss weight to~1, otherwise it is set to~0.}
\label{table:cifar_results}
\end{table*}

\newpage

\bibliography{main}
\bibliographystyle{plainnat}

\end{document}